\documentclass[letterpaper]{article} 
\usepackage{aaai2026}  
\usepackage{times}  
\usepackage{helvet}  
\usepackage{courier}  
\usepackage[hyphens]{url}  
\usepackage{graphicx} 
\usepackage{natbib}  
\usepackage{caption} 
\usepackage{algorithm}
\usepackage{algorithmic}

\usepackage{newfloat}
\usepackage{listings}
\DeclareCaptionStyle{ruled}{labelfont=normalfont,labelsep=colon,strut=off} 
\floatstyle{ruled}
\newfloat{listing}{tb}{lst}{}
\floatname{listing}{Listing}
\usepackage{amsmath,amsfonts}
\usepackage{array}
\usepackage{subfig}
\usepackage{textcomp}
\usepackage{overpic}
\usepackage{verbatim}
\usepackage{multirow}
\usepackage[table,xcdraw]{xcolor}\usepackage{colortbl}

\title{Dense Cross-Scale Image Alignment With Fully Spatial Correlation and Just Noticeable Difference Guidance}

\author{
    Jinkun You\textsuperscript{\rm 1}, Jiaxue Li\textsuperscript{\rm 2}, Jie Zhang\textsuperscript{\rm 1}, Yicong Zhou\textsuperscript{\rm 1}\thanks{Corresponding author.}
}
\affiliations {
    \textsuperscript{\rm 1}Department of Computer and Information Science, University of Macau\\
    \textsuperscript{\rm 2}School of Artificial Intelligence, China University of Geosciences (Beijing)\\
    youjinkun09@gmail.com, lijiaxue7@gmail.com, 
    jiezh1997@gmail.com, 
    yicongzhou@um.edu.mo
}

\usepackage{bibentry}

\begin{document}

\maketitle

\begin{abstract}
Existing unsupervised image alignment methods exhibit limited accuracy and high computational complexity. To address these challenges, we propose a dense cross-scale image alignment model. It takes into account the correlations between cross-scale features to decrease the alignment difficulty. Our model supports flexible trade-offs between accuracy and efficiency by adjusting the number of scales utilized. Additionally, we introduce a fully spatial correlation module to further improve accuracy while maintaining low computational costs. We incorporate the just noticeable difference to encourage our model to focus on image regions more sensitive to distortions, eliminating noticeable alignment errors. Extensive quantitative and qualitative experiments demonstrate that our method surpasses state-of-the-art approaches.
\end{abstract}


\section{Introduction}
Image alignment has received increasing attention in recent years \cite{jiang2022towards, kim2024omnistitch, li2024seam}. This task involves spatially registering images of the same scene captured from different viewpoints, aiming to maximize the consistency of overlapping regions. It forms the foundation for various applications, including deep camera calibration \cite{liao2023deep}, surround-view imaging systems \cite{kumar2023surround}, and multimodal image processing \cite{zhao2023cddfuse,you5372220hffnet}. This growing importance has driven efforts to optimize alignment accuracy and efficiency for practical deployment.

Traditional alignment methods rely on hand-crafted features to identify matched points between input images \cite{li2022automatic, li2019robust, lin2016seagull}. They establish transformations based on the point correspondences. Early approaches employ a single homography to achieve satisfactory results when the images share a common plane or differ only in camera rotation \cite{li2024automatic}. However, their performance degrades under parallax conditions. To overcome this limitation, non-linear local deformations were introduced to offer greater flexibility by predicting a mesh instead of a single homography. The thin-plate spline transformation has been adopted to enhance the alignment performance. Higher accuracy leads to greater content similarity in the overlapping regions of the aligned images. Nevertheless, focusing solely on accuracy can lead to projective distortion in non-overlapping areas. It causes straight lines to bend and objects to be over-stretched. Recent advances have incorporated geometric constraints to better preserve object shapes \cite{jia2021leveraging}.

While traditional methods yield promising results, they struggle with low-texture and large-parallax cases  \cite{nie2023parallax}. Deep networks have been utilized to address these limitations since they have powerful feature extraction and representation abilities \cite{deng2024enable,you2024two,ou2024clib}. They extract feature maps to generate a correlation tensor that encodes the matching relationships between images. This tensor is used to predict position offsets between image pixels to do image warping effectively \cite{detone2016deep}. Unsupervised methods are preferred recently since they do not need the ground truth offsets to train deep networks and are more suitable for real-world image pairs \cite{nie2021unsupervised, jia2023learning}. Early unsupervised methods estimate a single homography by maximizing content similarity in the overlapping regions of aligned images \cite{nguyen2018unsupervised}. To better handle parallax problems, the non-linear local alignment strategy is integrated in a coarse-to-fine manner \cite{nie2022depth}. Global offsets are first estimated from global features to maintain global consistency. Next, local offsets are computed from fine features for precise alignment. The global and local offsets are integrated to obtain the mesh used for image warping. Shape preservation is enforced by constraining the magnitudes and directions of the mesh.

Existing unsupervised methods have achieved impressive alignment accuracy, but they still face three key limitations. First, they do not explicitly account for scale variations. The apparent size of the same object can differ across images due to parallax, complicating the alignment process. Current approaches typically predict the transformation parameters using features of identical spatial dimensions. These intra-scale features cannot effectively capture scale discrepancies. Second, widely used correlation calculation methods are either ineffective or inefficient. The correlation layer is efficient but destroys spatial information \cite{rocco2017conv}. The contextual correlation layer and cost volume capture long-range correlation information to achieve better performance at the cost of high FLOPs and runtime, respectively \cite{nie2022depth}. Their advantages diminish when other modules also capture context information. Third, prevailing methods treat all image pixels equally during optimization. Not all pixels exhibit the same tolerance to distortion \cite{jiang2022toward}. Alignment errors are more noticeable to the human visual system when they occur in pixels that are more sensitive to distortion. As a result, the content similarity is more easily reduced for the overlapping areas of the aligned images.

To tackle these challenges, we introduce an unsupervised image alignment method with the following contributions:
\begin{itemize}
    \item We design a dense cross-scale image alignment model that leverages cross-scale features to enhance alignment accuracy. It allows for flexible trade-offs between effectiveness and efficiency based on user requirements.
    \item We introduce a fully spatial correlation module that flexibly utilizes the spatial information of the input features to further increase alignment accuracy. It can also maintain low computational costs.
    \item The just noticeable distortion estimation is used to guide image alignment. It helps an alignment model prioritize pixels with lower distortion tolerance, thus further increasing alignment accuracy.
    \item Extensive ablation studies and comparisons are conducted to validate the effectiveness and efficiency of the proposed method.
\end{itemize}

\section{Proposed Method}

\begin{figure*}[t]
  \centering
  \includegraphics[width=0.96\linewidth]{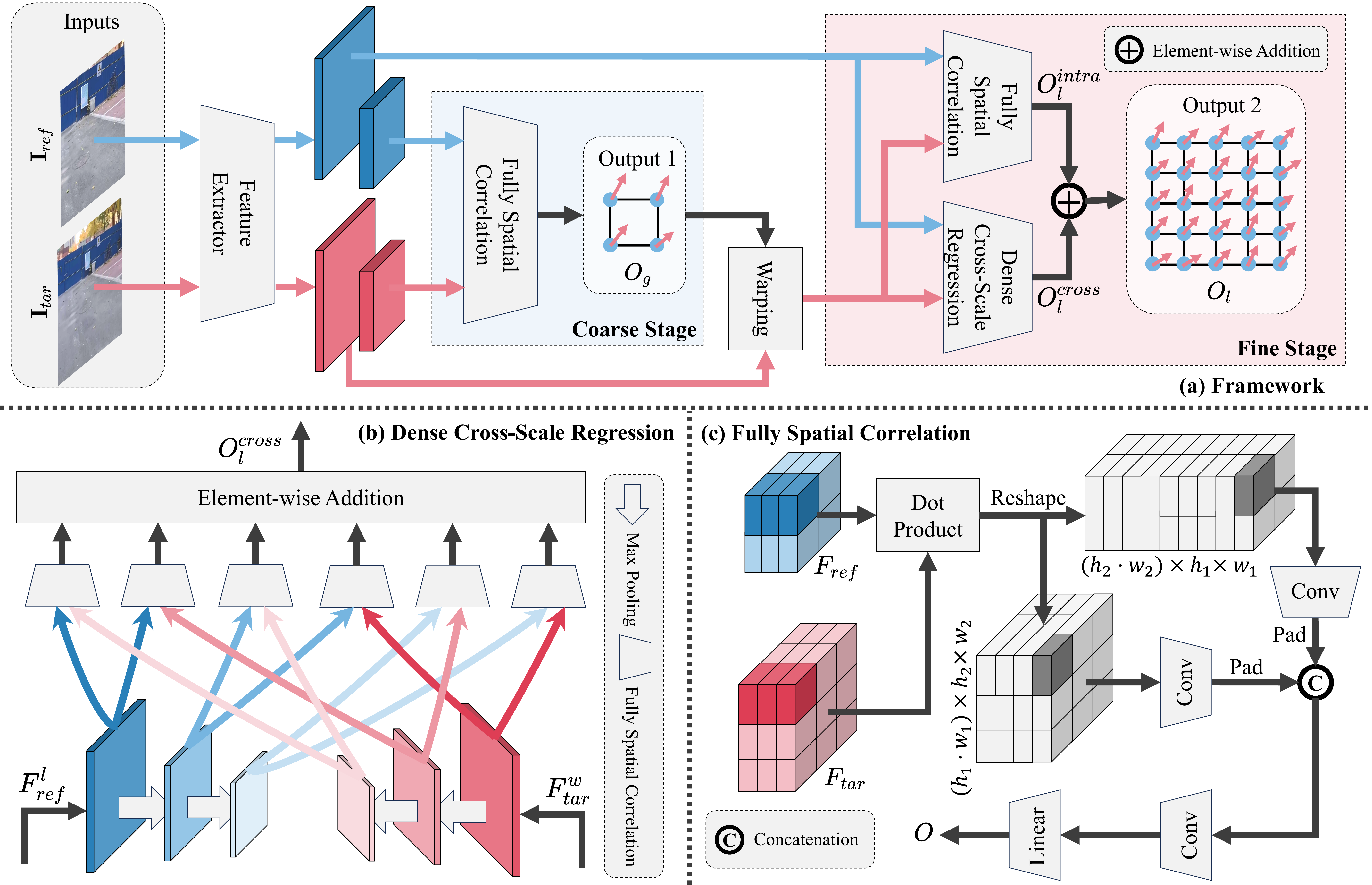}
  \caption{Illustration of the proposed alignment network. (a) The overall framework. (b) The proposed cross-scale regression. (c) The proposed fully spatial correlation. Our method uses a coarse-to-fine strategy to estimate the offsets of a mesh. The dense cross-scale regression module integrates the cross-scale information into the local offsets. The fully spatial correlation module utilizes the spatial information of both input features. The dot product is performed between the dark blue and dark red feature vectors to obtain the element in dark gray.}
  \label{fig:overview}
\end{figure*}

\subsection{Overall Structure}
The non-linear local deformation is employed to warp the target image $I_{tar}$ for the alignment with the reference image $I_{ref}$. It is implemented via a mesh $M^f\in\mathbb{R}^{U\times V\times 2}$, which has a spatial size of $U\times V$ and contains the horizontal and vertical positions. This alignment can be formulated as the optimization problem below:
\begin{equation}
    {M}^f_{opt}=\min_{{M}^f} \mathrm{P}({I}_{ref}, \mathrm{Warp}({I}_{tar}, {M}^f)),
    \label{eq:objective}
\end{equation}
where $\mathrm{P}(\cdot, \cdot)$ evaluates the alignment error and $\mathrm{Warp}(\cdot,\cdot)$ performs differentiable image warping.

The dense cross-scale alignment network is introduced to address the optimization problem more effectively. Instead of regressing the mesh directly, it predicts position offsets. As illustrated in $\textrm{Figure \ref{fig:overview}(a)}$, the network employs a coarse-to-fine strategy. Specifically, it extracts fine and coarse feature maps from both reference and target images to regress global offsets ${O}_{g}\in\mathbb{R}^{4\times 2}$ and local offsets ${O}_{l}\in\mathbb{R}^{U\times V\times 2}$, respectively. The global offsets $O_g$ are used to apply a homography transformation on a regular mesh $M\in\mathbb{R}^{U\times V\times 2}$ to ensure global consistency. Subsequently, $O_l$ is added to the transformation result to generate $M^f$ for fine-grained warping. Our method leverages cross-scale information from fine feature maps to enhance local deformation. A fully spatial correlation module is designed to assist in predicting both global and local offsets. Additionally, just noticeable difference (JND) guidance is incorporated to reduce the alignment errors in visually sensitive regions.

\subsection{Dense Cross-Scale Regression}
Large parallax results in scale variations for the same object across the reference and target images. Existing methods rely solely on intra-scale features and fail to accurately establish their relationships. This limits alignment performance and motivates the development of the dense cross-scale regression module to alleviate alignment difficulty. The cross-scale information is incorporated into the local offset $O_l$, which is expressed as
\begin{equation}
    O_l=O_l^{intra}+O_l^{cross},
\end{equation}
where $O_l^{intra}$ and $O_l^{cross}\in\mathbb{R}^{U\times V\times 2}$ are the offsets calculated from the intra-scale and cross-scale features, respectively. Notably, intra-scale information is integrated at both coarse and fine stages. The proposed module predicts $O_l^{cross}$ in a residual manner by leveraging diverse cross-scale information captured from the fine feature maps.

$\textrm{Figure \ref{fig:overview}(b)}$ illustrates the structure of the dense cross-scale regression module. Let ${F}^l_{ref}$ and ${F}^l_{tar}\in\mathbb{R}^{c\times h\times w}$ denote the fine features extracted from the reference and target images, respectively. The homography transformation is applied to ${F}^l_{tar}$ using the global offset $O_g$ to obtain the pre-aligned version ${F}^w_{tar}$. The cross-scale regression module takes ${F}^l_{ref}$ and ${F}^w_{tar}$ as inputs and first downsamples them. This downsampling operation is implemented using max pooling for efficiency while preserving important information. This process is repeated $N$ times, reducing the spatial size by half with each iteration. The results are expressed as:
\begin{equation}
    \begin{split}
        {F}^l_{i}=&\mathrm{MaxPool}_i(...\mathrm{MaxPool}_1({F}^l_{ref})...),\\
        {F}^w_{i}=&\mathrm{MaxPool}_i(...\mathrm{MaxPool}_1({F}^w_{tar})...),
    \end{split}
\end{equation}
where $\mathrm{MaxPool}_i(\cdot)$ represents the $i$-th max pooling operation and $i$ belongs to the positive integer set $\{1, 2, ..., N\}$; ${F}^l_{i}$ and ${F}^w_{i}$ are of size $\mathbb{R}^{c\times\frac{h}{2^i}\times\frac{w}{2^i}}$. For simplicity, we set ${F}^l_{0}$ and ${F}^w_{0}$ to ${F}^l_{ref}$ and ${F}^w_{tar}$, respectively. The cross-scale regression module establishes dense cross-scale relationships between the above features of the reference and target images. In this way, it fully utilizes the cross-scale information to enhance alignment accuracy. Specifically, a set of offsets is calculated based on ${F}^l_m$ and ${F}^w_n$ for $\forall m,n\in\{0, 1,...,N\}$ and $m\ne n$. In this way, we can obtain $N^2+N$ outputs. These offsets are aggregated to obtain $O_l^{cross}$. This process can be expressed as
\begin{equation}
    O_l^{cross}=\sum_{m,n\in[N]\textrm{ and }m\ne n}\mathrm{Net}({F}^l_m,{F}^w_n),
    \label{eq:aggregation}
\end{equation}
where $\mathrm{Net}(\cdot, \cdot)$ regresses local offsets from the input features and is implemented by the proposed fully spatial correlation module; $[N]$ denotes the set of natural numbers $\{0,1,...,N\}$. Only intra-scale features are used for alignment when $N$ is 0.

\subsection{Fully Spatial Correlation}
Let ${F}_{ref}\in\mathbb{R}^{c\times h_1\times w_1}$ and ${F}_{tar}\in\mathbb{R}^{c\times h_2\times w_2}$ represent the feature maps extracted from the reference and target images, respectively. A correlation tensor is computed from ${F}_{ref}$ and ${F}_{tar}$ to regress offsets. The contextual correlation layer (CCL) constructs this tensor by decomposing ${F}_{tar}$ into $h_2\cdot w_2$ overlapping kernels and convolving them with ${F}_{ref}$. The cost volume (CV) performs a Hadamard product on ${F}_{tar}$ and ${F}_{ref}$ using a sliding window approach. Both methods capture contextual information. However, the dense cross-scale regression module also establishes long-range relationships. It diminishes the advantages of the CCL and CV. Moreover, the large value of $h_2\cdot w_2$ results in numerous convolution operations, and the Hadamard product is computed repeatedly. They lead to high FLOPs and long runtime, respectively. Our dense cross-scale regression module requires $N^2+N$ such operations. The computational overhead is significant when the CCL and CV are employed.

To further improve effectiveness while maintaining low computational costs, we propose the fully spatial correlation module. $\textrm{Figure \ref{fig:overview}(c)}$ illustrates its architecture. Specifically, the module computes a 4D correlation tensor ${T}\in\mathbb{R}^{h_2\times w_2\times h_1\times w_1}$ to encode pairwise correlations through dot products between the feature vectors in 
${F}_{ref}$ and ${F}_{tar}$. This tensor ${T}$ must be processed through several convolutional and linear layers to regress offsets. To facilitate this, the correlation layer reshapes the 4D tensor ${T}$ into a 3D tensor ${T}_1\in\mathbb{R}^{(h2\cdot w2)\times h_1\times w_1}$. Nevertheless, this reshaping operation destroys the spatial information of ${F}_{tar}$. Convolutional layers can only utilize the spatial information of ${F}_{ref}$, limiting alignment accuracy. To fully leverage the spatial information from both feature maps, our module reshapes ${T}$ to construct another tensor ${T}_2\in\mathbb{R}^{(h_1\cdot w_1)\times h_2\times w_2}$. Both $T_1$ and $T_2$ are processed independently by convolutional layers for feature compression and refinement. Given the large dimensions of $h_1\cdot w_1$ and $h_2\cdot w_2$, we reduce the channel numbers to a more manageable size to enhance efficiency. The resulting outputs are then padded to match spatial dimensions for concatenation along the channel axis. The concatenation result is given by
\begin{equation}
    {T}^\prime=\mathrm{Cat}(\mathrm{Pad}(\mathrm{Conv}({T}_1)),
        \mathrm{Pad}(\mathrm{Conv}({T}_2))),
\end{equation}
where $\mathrm{Cat}(\cdot, \cdot)$, $\mathrm{Pad}(\cdot)$, and $\mathrm{Conv}(\cdot)$ represent the concatenation operation, padding operation, and convolution layers, respectively. Finally, ${T}^\prime$ undergoes further processing through a series of convolutional layers and linear layers to regress the offsets:
\begin{equation}
    {O}=\mathrm{Linear}(...(\mathrm{Linear}((\mathrm{Conv}(...\mathrm{Conv}({T}^\prime)...))...),
\end{equation}
where $\mathrm{Linear}(\cdot)$ denotes the linear layer. In this way, the proposed module can better utilize the spatial information from  ${F}_{ref}$ and ${F}_{tar}$ with low computational costs.

\begin{figure}[t]
  \centering
  \includegraphics[width=0.96\linewidth]{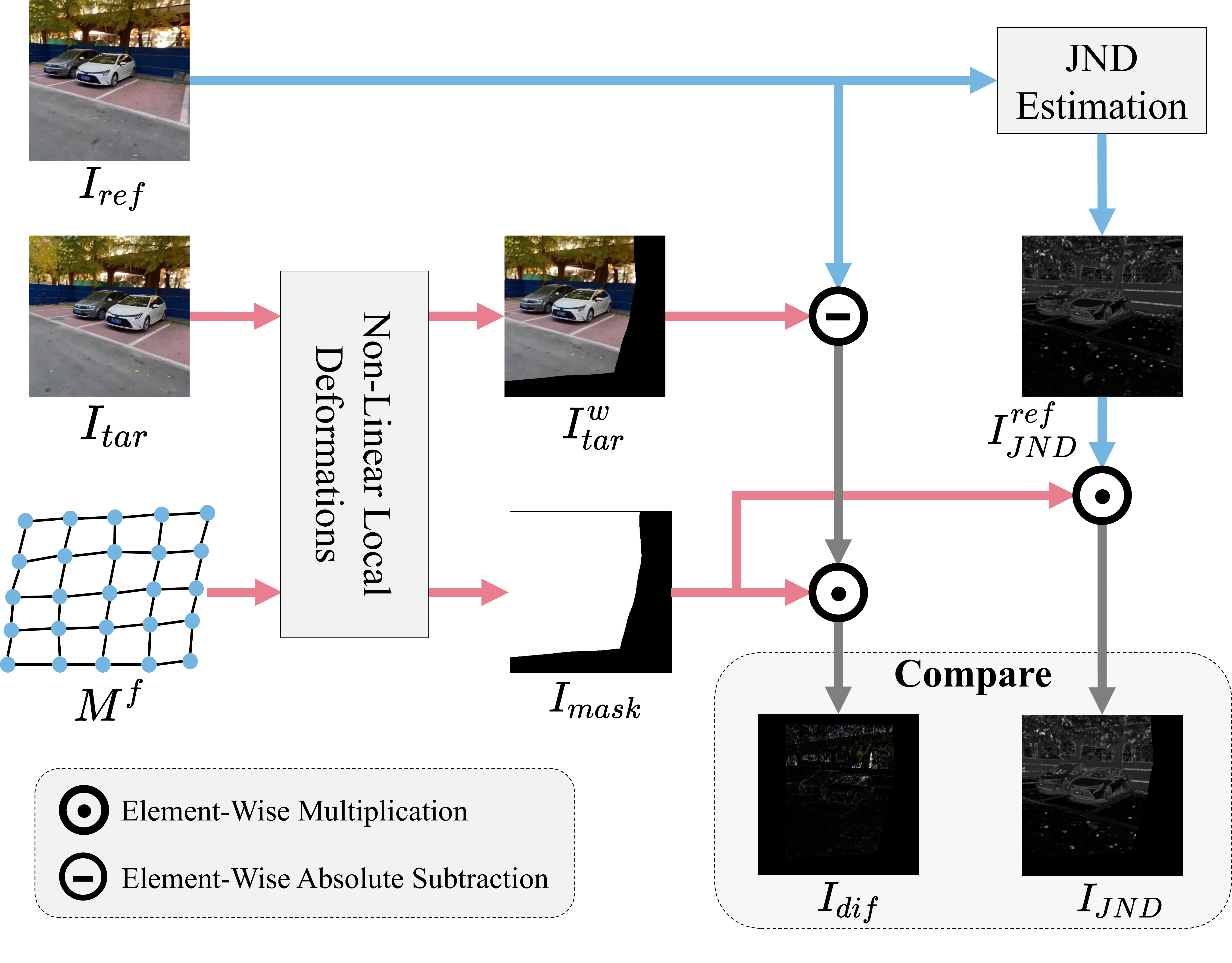}
  \caption{Illustration of the proposed JND guidance. The JND map is estimated for the reference image. The difference is calculated for the overlapping area to compare with the JND map to update network parameters.}
  \label{fig:jnd}
\end{figure}

\subsection{Just Noticeable Distortion Guidance}
As shown in $\textrm{Eq. (\ref{eq:objective})}$, an alignment model aims to find an optimal mesh that minimizes alignment errors. With the optimal mesh ${M}^f_{opt}$ and the one ${M}^f$ predicted by a model, we obtain the corresponding warped target images as follows:
\begin{equation}
    {I}^w_{opt}=\mathrm{Warp}({I}_{tar}, {M}^f_{opt}),
\end{equation}
\begin{equation}
    {I}^w_{tar}=\mathrm{Warp}({I}_{tar}, {M}^f)={I}^w_{opt}+{I_n},
\end{equation}
where ${I_n}$ denotes the noise introduced by ${M}^f$. Consequently, the optimization problem in $\textrm{Eq. (\ref{eq:objective})}$ can be reframed as minimizing ${I_n}$. Alignment accuracy is quantified by the similarity between the overlapping regions of the reference and warped target images. The noise $I_n$ degrades ${I}^w_{opt}$ to reduce this similarity, leading to alignment errors. However, the impact of $I_n$ on similarity varies across different instances, even at the same noise level, due to the characteristics of the human visual system \cite{jiang2022toward}. Distortions are more readily perceived in certain regions of an image, resulting in greater degradation of similarity in those areas. Inspired by this insight, we leverage the JND to further enhance alignment accuracy. The JND serves as an indicator of distortion tolerance for each pixel. A pixel change is imperceptible when it falls below the threshold value specified by the JND. By utilizing these threshold values, we encourage our method to focus on image regions that are more sensitive to distortions. $\textrm{Figure \ref{fig:jnd}}$ illustrates the proposed JND guidance strategy. It aims to ensure that distortions in the overlapping areas between the reference and warped target images remain below the JND thresholds.

We adopt the methodology from \cite{wu2015enhanced} to estimate the JND map ${I}^{ref}_{JND}$ for the reference image. Detailed estimation procedures are provided in the subsection ``JND Estimation'' of the supplementary material. Next, we calculate the difference between the reference image ${I}_{ref}$ and the warped target image ${I}^{w}_{tar}$ within the overlapping areas. A pixel's distortion is considered for parameter updates only if it exceeds the corresponding threshold in ${I}^{ref}_{JND}$. This consideration can be expressed as:
\begin{equation}
    I(x)=
    \begin{cases}
        0& \textrm{if }{I}_{dif}(x)\leq{I}_{JND}(x)\\
        {I}_{dif}(x)-{I}_{JND}(x)& \textrm{if }{I}_{dif}(x)>{I}_{JND}(x)
    \end{cases},
\end{equation}
\begin{equation}
    {I}_{dif} = |{I}_{ref}-{I}^{w}_{tar}|\odot{I}_{mask},
\end{equation}
\begin{equation}
    {I}_{JND}={I}^{ref}_{JND}\odot{I}_{mask}, 
\end{equation}
where $x$ denotes a pixel position; $\odot$ and $|\cdot|$ represent the Hadamard product and absolute value operation, respectively; ${I}_{mask}$ is a binary mask indicating overlapping and non-overlapping areas with values of one and zero, respectively. When updating the parameters of the alignment model, our goal is to minimize each element in $I$ as much as possible. We incorporate $I$ into the loss function to guide the parameter update. The loss function of the JND is defined as
\begin{equation}
    L_{JND}=\mathrm{Mean}(\mathrm{ReLU}({I}_{dif}-{I}_{JND})),
    \label{eq:jnd_loss}
\end{equation}
where $\mathrm{Mean}(\cdot)$ and $\mathrm{ReLU}(\cdot)$ represent the average operation and rectified linear unit, respectively.

\begin{table*}[!htbp]
\centering
\begin{tabular}{c|cccc|cccc|c|c}
\hline
                         & \multicolumn{4}{c|}{PSNR$\uparrow$}          & \multicolumn{4}{c|}{SSIM$\uparrow$}            &                       &                           \\ \cline{2-9}
\multirow{-2}{*}{Method} & Easy  & Moderate & Hard  & Average & Easy   & Moderate & Hard   & Average & \multirow{-2}{*}{FLOPs$\downarrow$}  & \multirow{-2}{*}{Time$\downarrow$} \\ 
\hline
$N=0$                      
& 30.46 & 26.04    & 21.74 & 25.63   & 0.9363 & 0.8805   & 0.7453 & 0.8426  & 443.1 G                                & 0.0174                    \\
$N=1$                      
& 30.88 & 26.58    & 22.26 & 26.13   & 0.9405 & 0.8915   & 0.7653 & 0.8552  & 527.1 G                               & 0.0246                    \\
$N=2$                      
& 30.95 & 26.62    & 22.29 & 26.18   & 0.9408 & 0.8918   & 0.7657 & 0.8556  & 562.9 G                               & 0.0305                    \\
$N=2$ w/ CCL                
& 30.72 & 26.48    & 22.16 & 26.01   & 0.9381 & 0.8882   & 0.7588 & 0.8509  & 1396.8 G                                & 0.0315                    \\ 
$N=2$ w/ CL              
&30.80  &26.46     &22.10  &26.01    &0.9394   &0.8887    &0.7591  &0.8516  & 556.3 G                              & 0.0285                    \\
$N=2$ w/ CV                
&\rule[0.05cm]{0.5cm}{0.5pt} &\rule[0.05cm]{0.5cm}{0.5pt}     &\rule[0.05cm]{0.5cm}{0.5pt}  &\rule[0.05cm]{0.5cm}{0.5pt}    &\rule[0.05cm]{0.5cm}{0.5pt}   &\rule[0.05cm]{0.5cm}{0.5pt}    &\rule[0.05cm]{0.5cm}{0.5pt}  &\rule[0.05cm]{0.5cm}{0.5pt}  & 431.9 G                                & 0.5580                    \\ 
$N=2$ w/o JND              
& 30.82 & 26.50    & 22.12 & 26.03   & 0.9403 & 0.8895   & 0.7599 & 0.8524  & 562.9 G                              & 0.0305                    \\
\hline
\end{tabular}
\caption{Ablation studies on each proposed module. The fully spatial correlation and JND guidance are employed by default. $N$ denotes the number of scales used to predict local offsets. CCL, CL, and CV represent the contextual correlation layer, correlation layer, and cost volume, respectively. From ``Easy'' to ``Hard'', the parallax becomes larger.}
\label{tab:ablation}
\end{table*}

\subsection{Loss Function}
The loss function is designed to optimize three critical objectives in image alignment tasks. First, it enforces precise spatial correspondence by maximizing image similarity between overlapping regions of the reference and warped target images. Second, it maintains the visual plausibility of the warped target image through naturalness constraints that preserve realistic structures. Third, it incorporates perceptual regularization to ensure that the introduced distortions remain below the threshold of human visual perception as possible. These requirements are mathematically formalized through three complementary loss components as follows:
\begin{equation}
    L=L_{content}+\alpha L_{shape}+\beta L_{JND},
\end{equation}
where $L_{content}$ and $L_{shape}$ are the content and shape preservation losses employed in \cite{nie2023parallax} and \cite{nie2022deep}; $L_{JND}$ is the distortion loss given in $\textrm{Eq. (\ref{eq:jnd_loss})}$. The details of  $L_{content}$ and $L_{shape}$ are shown in the subsection ``Loss Function'' of the supplementary material.

\begin{figure}[t]
    \centering
    \includegraphics[width=0.96\linewidth]{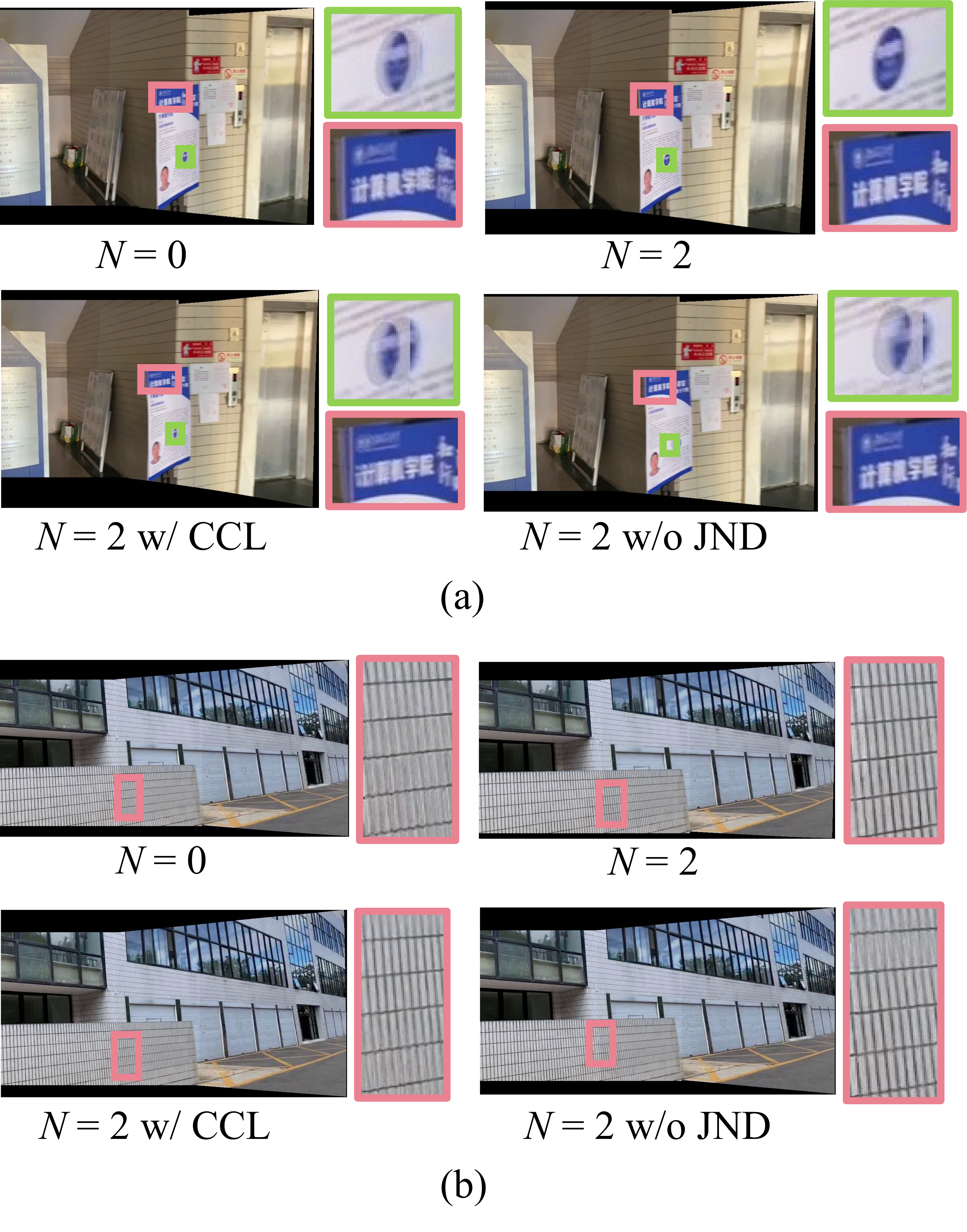}
    \caption{Ablation studies on our method. The red and green boxes zoom in on the regions with alignment errors. }
    \label{fig:abaltion_visual}
\end{figure}

\begin{figure*}[t]
\centering
\includegraphics[width=0.93\linewidth]{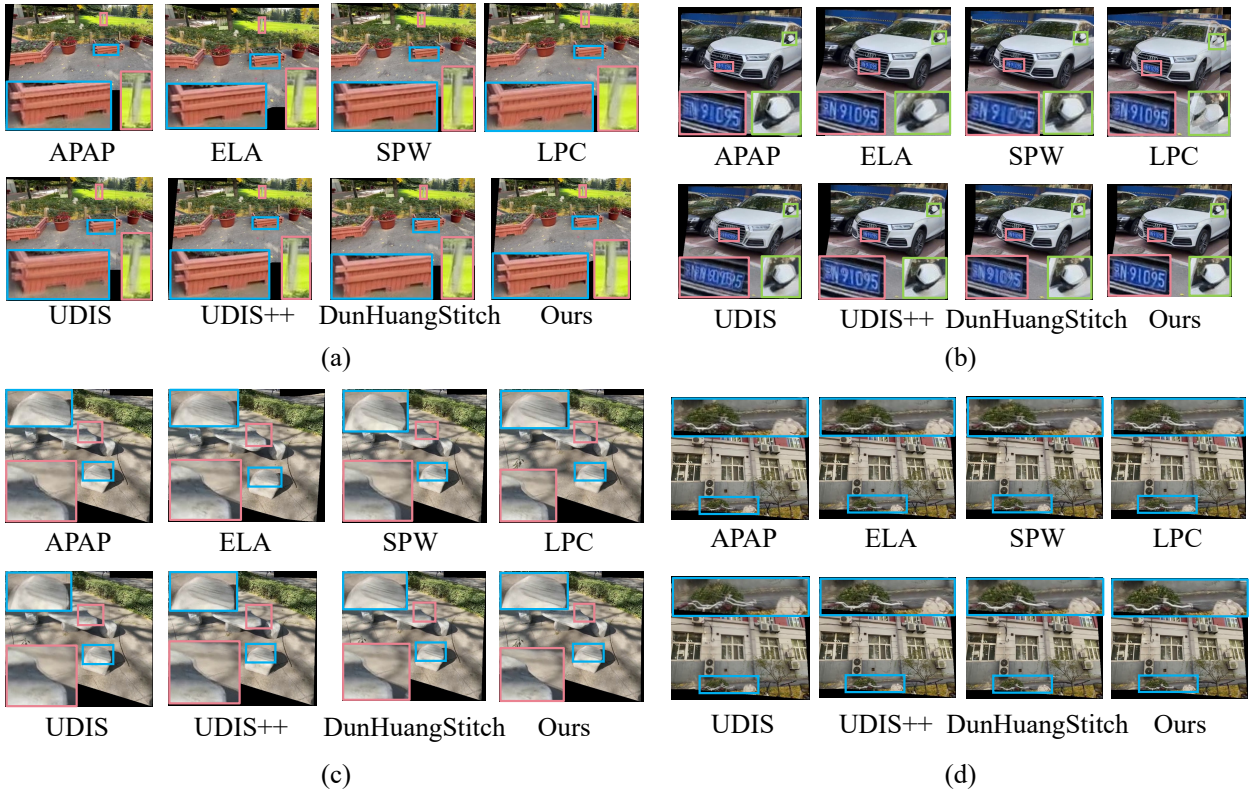}
\caption{Visual comparisons of different alignment methods on the UDIS-D dataset \cite{nie2021unsupervised}. The colorful boxes zoom in on the area with alignment errors.}
\label{fig:visual_comparison}
\end{figure*}

\section{Experiments}
\subsection{Experimental Settings}
\subsubsection{Datasets.} Following existing unsupervised methods, the UDIS-D dataset \cite{nie2021unsupervised} is used to train and assess our image alignment model. It is the only large-scale public dataset that contains real-world image pairs with parallax. No ground truth is provided. There are 10,440 image pairs for training and 1,106 pairs for testing, with each image sized $512\times512$. The small traditional dataset \cite{liao2020single} is also used to perform cross-dataset validation.

\subsubsection{Implementations.} Our alignment network is implemented in PyTorch and trained for 150 epochs on a single NVIDIA RTX A100 GPU. We use the Adam optimizer for parameter updates, with coefficients set to 0.9 and 0.999. The ResNet \cite{he2016deep} is employed as the feature extractor. The learning rate, batch size, mesh size, $\alpha$, $\beta$, and $N$ are set to ${10}^{-4}$, 4, $13\times13$, 10, 1, and 2, respectively.

\subsubsection{Metrics.} Due to the lack of ground truth, existing unsupervised methods measure the accuracy of image alignment by the similarity between overlapping regions of the reference and warped target images. The PSNR and SSIM are two commonly used metrics for this evaluation.

\begin{table*}[t]
\centering
{
\begin{tabular}{c|cccc|cccc}
\hline
                         & \multicolumn{4}{c|}{PSNR$\uparrow$}        & \multicolumn{4}{c}{SSIM$\uparrow$}         \\ \cline{2-9} 
\multirow{-2}{*}{Method} & Easy & Moderate & Hard & Average & Easy & Moderate & Hard & Average \\ \hline
$I_{3\times3}$                  
&15.87      &12.76      &10.68     &12.86      &0.530      &0.286      &0.146     &0.303    \\
APAP                   
&27.99      &24.29      &20.17     &23.74      &0.899      &0.832      &0.683     &0.792    \\
ELA                 
&29.74      &25.36      &19.47     &24.30     &0.923      &0.862      &0.704     &0.816    \\
SPW                  
&28.15      &23.39      &19.86     &23.32      &0.907      &0.799      &0.631     &0.764    \\
LPC                 
&27.26      &22.87      &19.38     &22.78      &0.883      &0.776      &0.623     &0.747    \\
IHN
&22.16      &18.17      &14.53    &17.90      &0.747      &0.558      &0.391     &0.547    \\
RHWF    
&20.10      &16.42      &13.51    &16.35        &0.688      &0.481     &0.331  &0.483  \\
MCNet    
&22.52      &18.68      &15.45     &18.53      &0.768      &0.591      &0.422     &0.576    \\
UDIS                 
&25.16      &20.96      &18.36     &21.17     &0.834      &0.669      &0.495     &0.648    \\
DAMG               
&29.52      &25.24      &21.20     &24.89     &0.923      &0.859      &0.708     &0.817    \\
UDIS++                   
&\underline{30.21}      &\underline{25.83}      &21.60     &\underline{25.43}     &\underline{0.934}      &\underline{0.876}      &0.739     &\underline{0.838}    \\
DunHuangStitch    
&29.47      &25.60      &\underline{22.01}     &25.31      &0.921      &0.859      &\underline{0.741}     &0.830    \\
Ours                
&\textbf{30.95}      &\textbf{26.62}      &\textbf{22.29}      &\textbf{26.18}      &\textbf{0.941}      &\textbf{0.892}      &\textbf{0.766}      &\textbf{0.856}    \\
\hline
\end{tabular}
}
\caption{Alignment accuracy comparisons of different methods on the UDIS-D dataset \cite{nie2021unsupervised}. The best and second-best results are written in bold and underlined, respectively. $N$ denotes the number of scales used by our method. $I_{3\times3}$ indicates that no alignment is performed. From ``Easy'' to ``Hard'', the parallax becomes larger.}
\label{tab:comparison_baslines}
\end{table*}

\subsection{Ablation Studies}
We conduct ablation studies to validate the effectiveness of each proposed module. $\textrm{Table \ref{tab:ablation}}$ shows the alignment accuracy and computational costs.

\subsubsection{Cross-Scale Regression.} 
Our method with $N=0$ only uses the intra-scale information for alignment, while those with $N>0$ incorporate the cross-scale information. Parallax causes the scale variation problem. As shown in $\textrm{Table \ref{tab:ablation}}$, alignment accuracy improves across different parallax scenarios as
$N$ increases. These results underscore the importance of the cross-scale information in addressing the variation problem. The computational costs also increase with increasing $N$. The additional cross-scale information reduces efficiency. Thus, our method allows for various trade-offs between accuracy and computational costs by adjusting $N$. Notably, the accuracy gains become marginal when increasing $N$ from 1 to 2. This leads us to set $N$ to 2.

\subsubsection{Fully Spatial Correlation.} 
According to $\textrm{Table \ref{tab:ablation}}$, replacing the fully spatial correlation module with the CCL and CL results in decreased alignment accuracy. One advantage of the CCL is its ability to capture contextual information through convolution operations. The dense cross-scale module can utilize cross-scale features to establish long-range relationships. This enables the CL to achieve performance comparable to the CCL. The fully spatial correlation module can better facilitate offset regression through its flexible spatial information integration ability. Thus, it outperforms the CCL and CL. Furthermore, our method with $N=1$ has higher alignment accuracy than the CCL and CL with $N=2$. In this case, the CCL and CL incur longer inference times and higher FLOPs. Especially, the CCL has approximately 2.5 times the FLOPs of our method. While parallel computation techniques can help reduce inference time, they do not alleviate the increased energy demands associated with higher FLOPs. Besides, the varying spatial sizes of cross-scale features render the CV unsuitable for our approach, as it necessitates input features with consistent spatial dimensions. Although the CV has lower FLOPs, it exhibits nearly 20 times the runtime compared to our method. Overall, our approach achieves superior alignment accuracy while maintaining low computational costs.

\subsubsection{JND Guidance.} The alignment accuracy is further enhanced under the guidance of the JND. It makes the alignment errors less perceptible in line with the human visual system. Besides, the JND guidance does not increase inference time and FLOPs, as it is used only for model training.

\subsubsection{Visualization.}  $\textrm{Figure \ref{fig:abaltion_visual}}$ presents the visualization results. To highlight alignment errors, we apply the average fusion to the reference and warped target images. In $\textrm{Figure \ref{fig:abaltion_visual}(a)}$, our method with $N=2$ exhibits superior alignment performance compared to the other three methods. These three methods introduce distortions in the text, board contour, and blue circle. In $\textrm{Figure \ref{fig:abaltion_visual}(b)}$, the method with $N=2$ does not produce obvious artifacts. In contrast, the alignment errors cause blurring: at the top for the method with the CCL, at the bottom for the method without JND guidance, and throughout the whole area for the method with $N=0$.

\begin{figure}[t]
    \centering
    \includegraphics[width=0.93\linewidth]{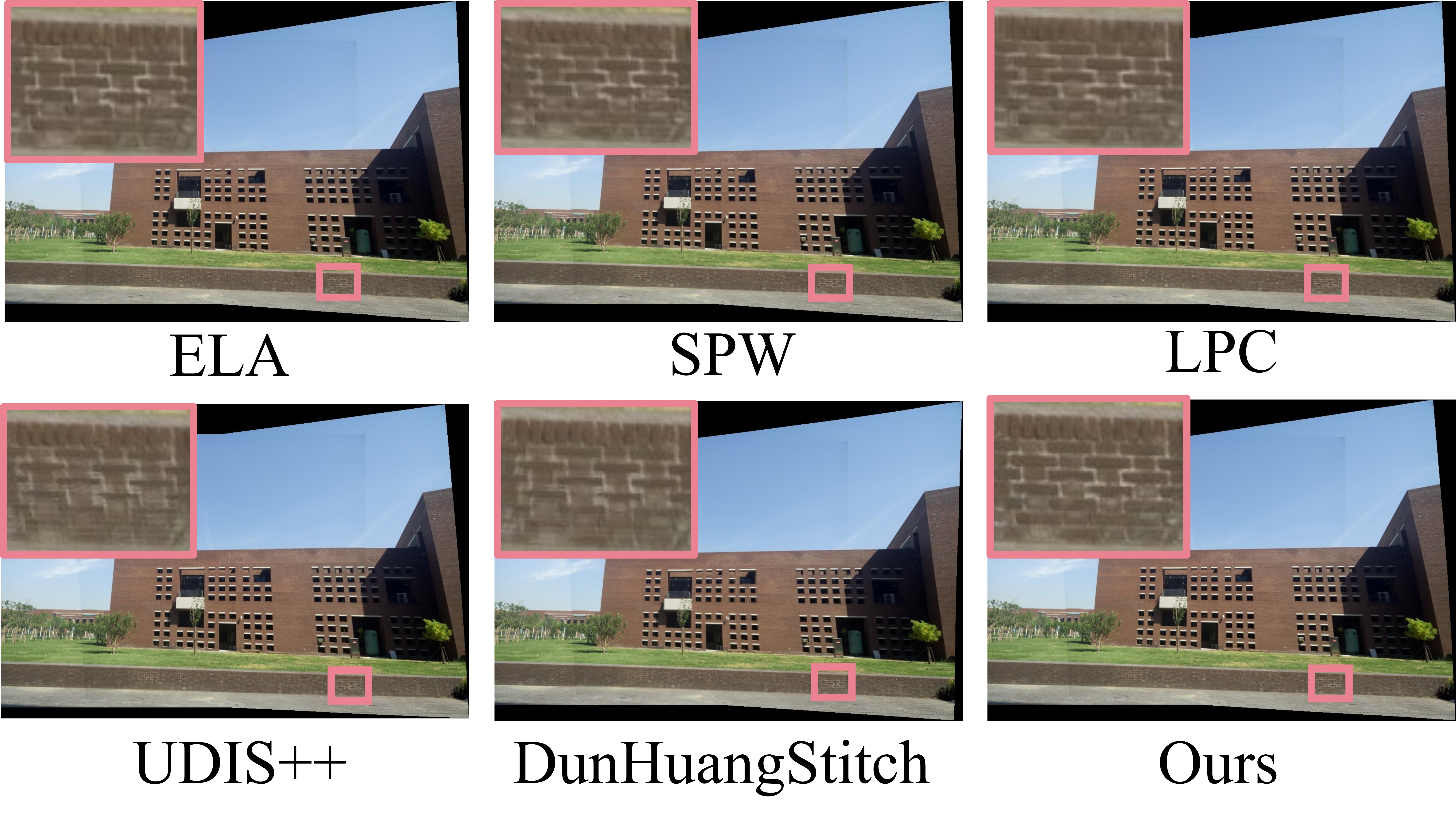}
    \caption{Validation results on the cross-dataset image ``19'' \cite{liao2020single}. The red box zooms in on the area with alignment errors.}
    \label{fig:visual_cross1}
\end{figure}

\subsection{Comparisons with Baselines}
Our method is compared with state-of-the-art unsupervised image alignment methods and homography estimation methods. The baselines include APAP \cite{zaragoza2013as}, ELA \cite{li2018parallax}, SPW \cite{liao2020single}, LPC \cite{jia2021leveraging}, UDIS \cite{nie2021unsupervised}, DAMG \cite{nie2022depth}, UDIS++ \cite{nie2023parallax}, DunHuangStitch \cite{mei2024dunhuangstitch}, IHN \cite{cao2022iterative}, RHWF \cite{cao2023recurrent} and MCNet \cite{zhu2024mcnet}.

\subsubsection{Quantitative Results.} 
$\textrm{Table \ref{tab:comparison_baslines}}$ reports the alignment accuracy of different methods. Compared to the baselines, our method improves the PSNR and SSIM values by at least 0.74/0.007, 0.79/0.016, and 0.28/0.025 for easy, moderate, and hard parallax scenarios, respectively. The average improvements in PSNR and SSIM are 0.75 and 0.018, respectively. Parallax causes the scale variation problem for the same object across the image pair. Our method outperforms all the baselines in various parallax scenarios. It can better address the scale variation problem.

\subsubsection{Qualitative Results.} $\textrm{Figure \ref{fig:visual_comparison}}$ displays the visualization results of different methods on the UDIS-D dataset. The average fusion is applied to highlight alignment errors. In $\textrm{Figure \ref{fig:visual_comparison}(a)}$, the planter outlined in blue shows misalignment across all competing methods. ELA, UDIS++, and DunhuangStitching exhibit less blurring at the top but more at the bottom. Other baselines demonstrate artifacts across the whole planter. The competing methods also cause noticeable distortions to the pole outlined in red. In $\textrm{Figure \ref{fig:visual_comparison}(b)}$, significant misalignment is observed in the license plate, except for ELA, LPC, and our method. However, ELA and LPC display artifacts in the side mirror. It is also observed from $\textrm{Figures \ref{fig:visual_comparison}(c) and \ref{fig:visual_comparison}(d)}$ that misalignment occurs in the areas of the stone table, the stone chair, and the handlebar crossbar for the competing methods. The scale of the stone chair and desk in $\textrm{Figure 4(c)}$ differs across the reference and target images. Our method achieves more plausible visual results and thus can better address the scale variation problem.

\subsubsection{Cross-Dataset Validation.} 
\begin{figure}[t]
    \centering
   \includegraphics[width=0.93\linewidth]{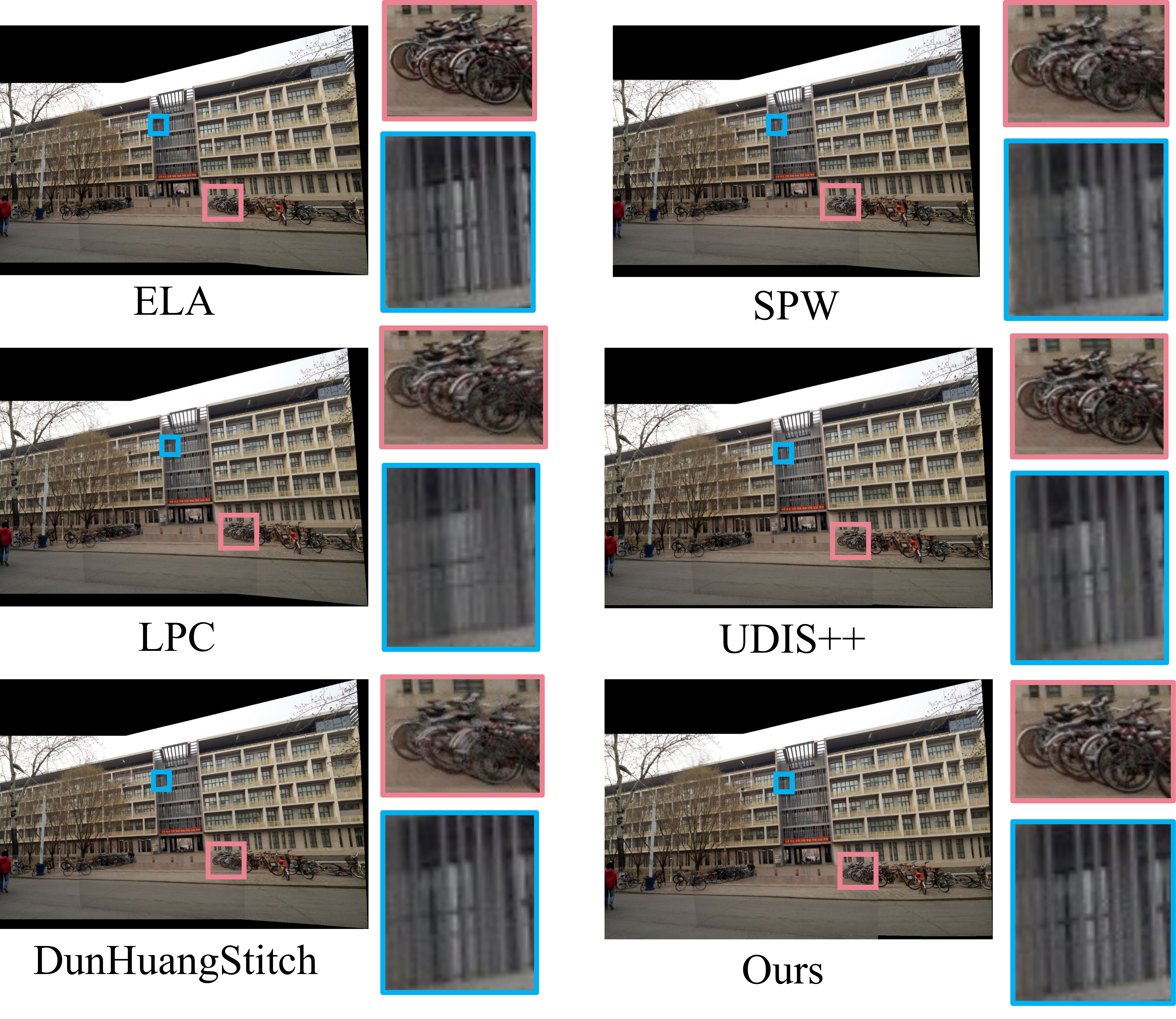}
    \caption{Validation results on the cross-dataset image "40" \cite{liao2020single}. The red and blue boxes zoom in on the area with alignment errors.}
    \label{fig:visual_cross2}
\end{figure}

We also evaluate the generalization of the proposed method on cross-dataset images \cite{liao2020single}. Following \cite{nie2023parallax}, the iterative optimization strategy is adopted. $\textrm{Figures \ref{fig:visual_cross1} and \ref{fig:visual_cross2}}$ present the visualization results. In $\textrm{Figure \ref{fig:visual_cross1}}$, noticeable artifacts appear at the top for ELA, SPW, and LPC, as well as at the bottom for UDIS++ and DunHuangStitch. $\textrm{Figure \ref{fig:visual_cross2}}$ reveals evident misalignment in areas with bicycles for SPW, LPC, and DunhuangStitch. All competing methods exhibit significant artifacts on the left side of the window fence. We also perform user studies to compare our method with UDIS++ and DunHuangStitch. Details are provided in the subsection ``Cross-Dataset Validation'' of the supplementary material. Feedback indicates that our method produces more visually plausible results than the two competing approaches.

\subsubsection{Runtime.} We show the runtime for images of varying resolutions in the subsection ``Runtime Comparison'' of the
supplementary material. Our method is competitive.

\section{Conclusion}
In this paper, we introduced a dense cross-scale image alignment method. It captured the cross-scale features to increase image alignment. The number of scales can be adjusted to balance accuracy and computational costs. A fully spatial module was designed to further increase accuracy while maintaining low computational costs. We also exploited the JND to encourage the alignment model to focus on regions sensitive to distortions, thus improving performance. Ablation studies validated the effectiveness of the proposed modules. Extensive experiments were conducted to demonstrate the superiority of the proposed method over the state-of-the-art approaches in both effectiveness and efficiency.

\section*{Acknowledgements}
This work was funded by the Science and Technology Development Fund, Macau SAR (File no. 0049/2022/A1, 0050/2024/AGJ), by the University of Macau (File no. MYRG2022-00072-FST, MYRG-GRG2024-00181-FST).

\bibliography{aaai2026}

\end{document}